\documentclass{article}

\usepackage[preprint]{neurips_2023}

\usepackage[utf8]{inputenc} 
\usepackage[T1]{fontenc}    
\usepackage{hyperref}       
\usepackage{url}            
\usepackage{booktabs}       
\usepackage{amsfonts}       
\usepackage{nicefrac}       
\usepackage{microtype}      
\usepackage{xcolor}         
\usepackage{wrapfig}
\usepackage{subfigure}
\newcommand{\ignore}[1]{{}}

\title{XRoute Environment: A Novel Reinforcement Learning Environment for Routing}

\author{%
  Zhanwen Zhou
  \\
  School of Computer Science\\
  Sun Yat-sen University\\
  Guangzhou 510006, China \\
  \texttt{zhouzhw26@mail2.sysu.edu.cn} \\
  \And
  Hankz Hankui Zhuo \\
  School of Computer Science\\
  Sun Yat-sen University\\
  Guangzhou 510006, China \\
  \texttt{zhuohank@mail.sysu.edu.cn} \\
  \AND
  Xiaowu Zhang \\
  School of Computer Science\\
  Sun Yat-sen University\\
  Guangzhou 510006, China \\
  \texttt{zhangxw87@mail.sysu.edu.cn} \\
  \And
  Qiyuan Deng \\
  School of Computer Science\\
  Sun Yat-sen University\\
  Guangzhou 510006, China \\
  \texttt{dengqy26@mail.sysu.edu.cn} \\
}

\def\ours{\tt XRoute}

\newcommand{\tabincell}[2]{\begin{tabular}{@{}#1@{}}#2\end{tabular}}

\usepackage{amsmath}
\usepackage{listings}
\usepackage{multirow}
\usepackage[pdftex]{graphicx}
\usepackage{pythonhighlight}
\usepackage{tablefootnote}

\begin{document}

\maketitle

\begin{abstract}
Routing is a crucial and time-consuming stage in modern design automation flow for advanced technology nodes. Great progress in the field of reinforcement learning makes it possible to use those approaches to improve the routing quality and efficiency. However, the scale of the routing problems solved by reinforcement learning-based methods in recent studies is too small for these methods to be used in commercial EDA tools. We introduce the {\ours} Environment, a new reinforcement learning environment where agents are trained to select and route nets in an advanced, end-to-end routing framework. Novel algorithms and ideas can be quickly tested in a safe and reproducible manner in it. The resulting environment is challenging, easy to use, customize and add additional scenarios, and it is available under a permissive open-source license. In addition, it provides support for distributed deployment and multi-instance experiments. We propose two tasks for learning and build a full-chip test bed with routing benchmarks of various region sizes. We also pre-define several static routing regions with different pin density and number of nets for easier learning and testing. For net ordering task, we report baseline results for two widely used reinforcement learning algorithms (PPO and DQN) and one searching-based algorithm (TritonRoute). The {\ours} Environment will be available at \url{https://github.com/xplanlab/xroute\_env}.
\end{abstract}

\section{Introduction}

In very large-scale integration (VLSI) design, routing is challenging and has become a vital bottleneck in real world applications due to the complicated design rules and large solution space \citep{DBLP:journals/tcad/ChenPLY20}. In general, the routing task is separated into global routing and detailed routing. The former aims at partitioning the whole routing region into individual cells called GCell and routing nets on GCell according to the prediction of congestion \citep{DBLP:journals/tcad/LiuKLC13}, while the latter builds rectilinear wiring interconnects that are consistent with design rules \citep{DBLP:conf/ispd/ParkKKGLC19}. The detailed routing attempts to realize the segments and vias according to the global routing solution, and aims at minimizing design rule violations, wirelength and vias used.

\begin{figure*}[!ht]
    \centering
    \subfigure[Resource conflicts in routing]{
        \begin{minipage}[t]{0.38\textwidth}
        \centering
	    {\includegraphics[width=0.8\textwidth]{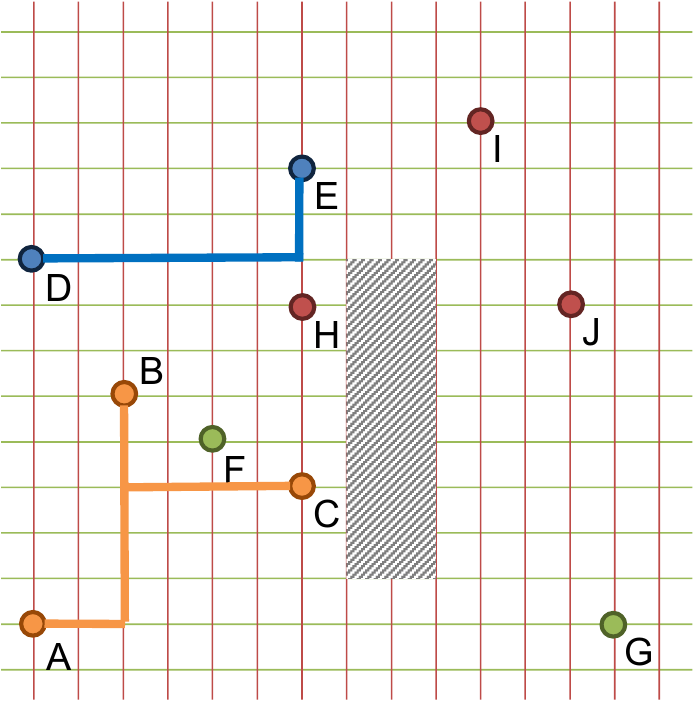}} 
	\end{minipage}
	\label{fig:net_order_conflicts1}
    }
    \subfigure[Ripup and reroute conflicted nets]{
        \begin{minipage}[t]{0.38\textwidth}
        \centering
   		\includegraphics[width=0.8\textwidth]{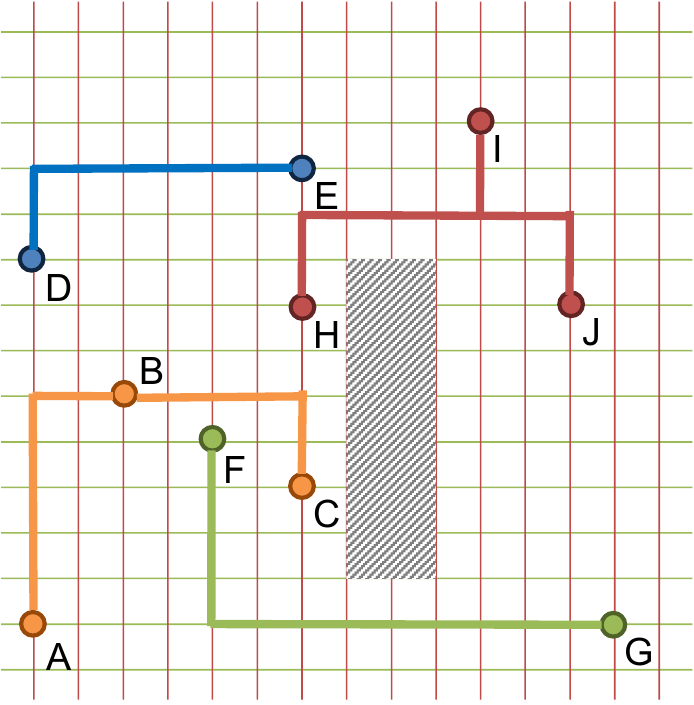}
        \end{minipage}
	\label{fig:net_order_conflicts2}
    }
    \subfigure{
	\begin{minipage}[t]{0.15\textwidth}
		{\includegraphics[width=0.9\textwidth]{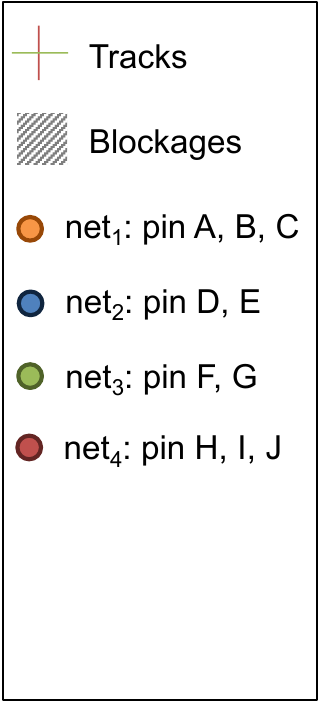}}
	\end{minipage}
    }
    \caption{Sequential routing is sensitive to the net order.}
    \label{fig:net_order_conflicts}
\end{figure*}

Maze-based and A*-based searching methods have enjoyed huge success in global and detailed routing. Until now, they are still used in many commercial electronic design automation (EDA) software. However, these methods tend to analyze and simulate some specific characteristics in routing, and hardcode them in the algorithm. They often explore a lot of unnecessary nodes during searching for the target. They seldom take into account the entire routing environment so as to bypass the obstacles as early as possible. What's worse, the current routing net based on these methods will not consider the routability of the following nets, which causes lots of rip-up and rerouting. As shown in Figure \ref{fig:net_order_conflicts1}, $net_1$ and $net_2$ are routed sequentially by maze routing, but these temporal optimal routed paths block the following connections of $net_3$ and $net_4$. The general solution for this situation is to ripup $net_1$ and $net_2$ and reroute $net_3$ and $net_4$ first as in Figure \ref{fig:net_order_conflicts2}. It shows that net ordering and net routing are crucial to VLSI routing.

The goal of reinforcement learning (RL) is to train smart agents that can interact with their environment and solve complex tasks \citep{Sutton2018}. Great progress in the field of deep reinforcement learning (DRL) makes it possible to use those methods to accelerate the routing process in net ordering and routing. However, the scale of the routing problems solved by the RL-based methods in recent studies is too small for those methods to be used in the commercial VLSI EDA tools. In this paper, we propose a novel RL environment, namely {\ours} Environment, standing for self-learning (denoted by \textbf{X}) for detailed routing (denoted by \textbf{Route}), to train agents to repeatedly recommend routing actions (representing ``net order'' or ``net edge'') in an end-to-end routing framework for detailed routing. Our {\ours} Environment is composed of three components, as shown below:
\begin{itemize}
    \item The first component is a highly-customized routing engine that simulates the detailed routing process. As far as we know, this is the first open-source reinforcement learning environment available for detailed routing learning.
    \item The second is a versatile set of benchmark tasks of varying size and difficulties that can be used to compare different algorithms, based on ISPD-2018 \citep{Mantik2018ISPD} and ISPD-2019 \citep{Liu2019ISPD} detailed routing testcases in our first version. We also pre-define several static regions with different pin density and number of nets for easier learning and testing. We evaluate two widely used algorithms on these benchmarks, providing a basic set of reference results for future comparison.
    \item The third is a dashboard that is used to exhibit the detailed routing results such as wirelength, vias used and design rule violations during model validation. Users can observe the detailed routing process and optimize their algorithms accordingly.
\end{itemize}
To help users utilizing our {\ours} Environment, we present two example tasks (net ordering and net routing) for learning to improve the quality and efficient of detailed routing. Users can follow the same way to define more RL tasks.

\section{Motivation and related work}
While there have been a variety of reinforcement learning algorithms for VLSI routing, there still exist open issues as shown bellow: 

\textbf{Lack of benchmarks from real chips.} 
Many routing problems proposed with their RL-based algorithms are generated by a self-developed generator. 
It lacks of benchmarks from real chips to verify the adaptability of the routing approaches in industrial advanced technology nodes. For example, DQN\_GlobalRouting \citep{LiaoGR2019} model the global routing problem as a Markov Decision Process (MDP) and uses deep Q-learning (DQN) to solve the routing problem by leveraging the conjoint optimization mechanism of deep reinforcement learning. A global routing problems generator is also developed to automatically generate parameterized problems with different scales and constraints. However, their nets are decomposed into a set of two-pin connections, and the environment is only used for learning to route from one source to one target. 
Alpha-PD-Router \citep{Gandhi2019} is a data-independent RL-based routing model, which learns to route a circuit and correct short violations. It is trained and tested on the $1 \times 5 \times 5$ routing grid with 3 two-pin nets. He et al. \citep{He2022} model circuit routing as a sequential decision problem and solve it in Monte Carlo tree search with DRL-guided rollout. In their environment, 100 randomly generated, routable single-layer circuits on a grid of $30 \times 30$ with less than 10 nets are used as the test bed for two-pin nets routing learning. Ju et al. \citep{Ju2021} propose a detailed router based on multi-agent reinforcement learning for handling conflicting nets, whose size far smaller than real chips.

\textbf{Not for detailed routing.} 
Circuit Training \citep{CircuitTraining2021} is an open-source framework created by Google Brain team for generating chip floor plans with distributed DRL \citep{mirhoseini_graph_2021}. It can place netlists with hundreds of macros and millions of standard cells. However, Circuit Training employ the half-perimeter wirelength (HPWL), the half-perimeter of the bounding box for all pins in a net, to approximate the wirelength of the net and do not run the real routing. 
PRNet \citep{ChengRuoyu2022} is an RL-based model for mixed-size macro placement. A one-shot conditional generative routing model is devised to perform one-shot global routing on sample grid graph with size of $64\times64$ or $128\times128$. REST \citep{DBLP:conf/dac/LiuCY21} is a reinforcement learning framework for rectilinear Steiner minimum tree construction. The model is trained with random generated point sets as pins from degree 3 to 50. Those approaches and their environments focus on placement or global routing, not on detailed routing.

\textbf{Not open-source.} 
There indeed are some RL-based detailed routing algorithms. For example, Lin et al. \citep{Lin2022} propose an asynchronous RL framework using Dr.CU \citep{Chen2020} as router to automatically search for optimal ordering strategies. The model is trained and tested on several dense clips with around 500 nets, which is selected from ISPD-2018 \citep{Mantik2018ISPD} and ISPD-2019 \citep{Liu2019ISPD} benchmarks. 
Ren and Fojtik \citep{Ren2021} applies genetic algorithm to create initial routing candidates and uses RL to fix the design rule violations incrementally on their developed program called Sticks. Unfortunately, there is no open-source license for those algorithms, which makes it hard to control over the environment dynamics and generate realistic evaluation scenarios to comprehensively test generalization in different chips with varying sizes and complexities.

\section{{\ours} Environment}


\subsection{Routing as a RL problem}

The background introduction of detailed routing is described in Appendix \ref{background_of_detailed_routing}.
Detailed Routing can be formulated as the MDP tuple $\langle \mathcal{S}, \mathcal{A}, \mathcal{P}, \mathcal{R}, \gamma \rangle$. Each state $s$ in the state space $\mathcal{S}$ consist of 3D grid graph routing environment information, netlist, pin position, routed paths for the current routing net. The action space $\mathcal{A}(s)$ is the set of available state-dependent actions, which represent the next net (or path) to be routed in the grid based on a given routing strategy. The transition model $\mathcal{P}(s^{\prime}|s,a)$ transitions to a new state $s^{\prime} \in \mathcal{S}$ given that action $a \in \mathcal{A}$ was executed from state $s$. The bound reward function $\mathcal{R}:\mathcal{S}\times\mathcal{A}\rightarrow\mathbb{R}$ is defined as the reward returned by the routing environment after routing a net (or path) in the grid. It is determined by the wirelength of the routed path, the number of vias used, and the design rule violations occurred after routing. $\gamma \in [0, 1]$ is the discount factor. A policy $\pi:\mathcal{S}\times\mathcal{A}\rightarrow[0, 1]$ is a mapping from a state to a distribution over actions where $\pi(a|s)$ denotes the probability of taking action $a$ in state $s$. The goal of detailed routing is to obtain a policy which maximizes the expected sum of discounted rewards ($\mathcal{J}_\pi^\mathcal{R}$ in Equation \ref{eq:objective}). That is to minimize the overall wirelength, via count and design rule violations of the routing solution on the basis of all nets are routed successfully.

\begin{equation} \label{eq:objective}
    \mathop{max}\limits_\pi \mathcal{J}_\pi^\mathcal{R}(s):=\mathop{\mathbb{E}}\limits_{a\sim\pi,s\sim\mathcal{P}}\left[\sum\limits_t\gamma^t\mathcal{R}(s_t, a_t)\rbrack\right]
\end{equation}

\subsection{Routing architecture}

\begin{wrapfigure}{r}{0.63\textwidth}
    \centering
    \includegraphics[width=0.6\textwidth]{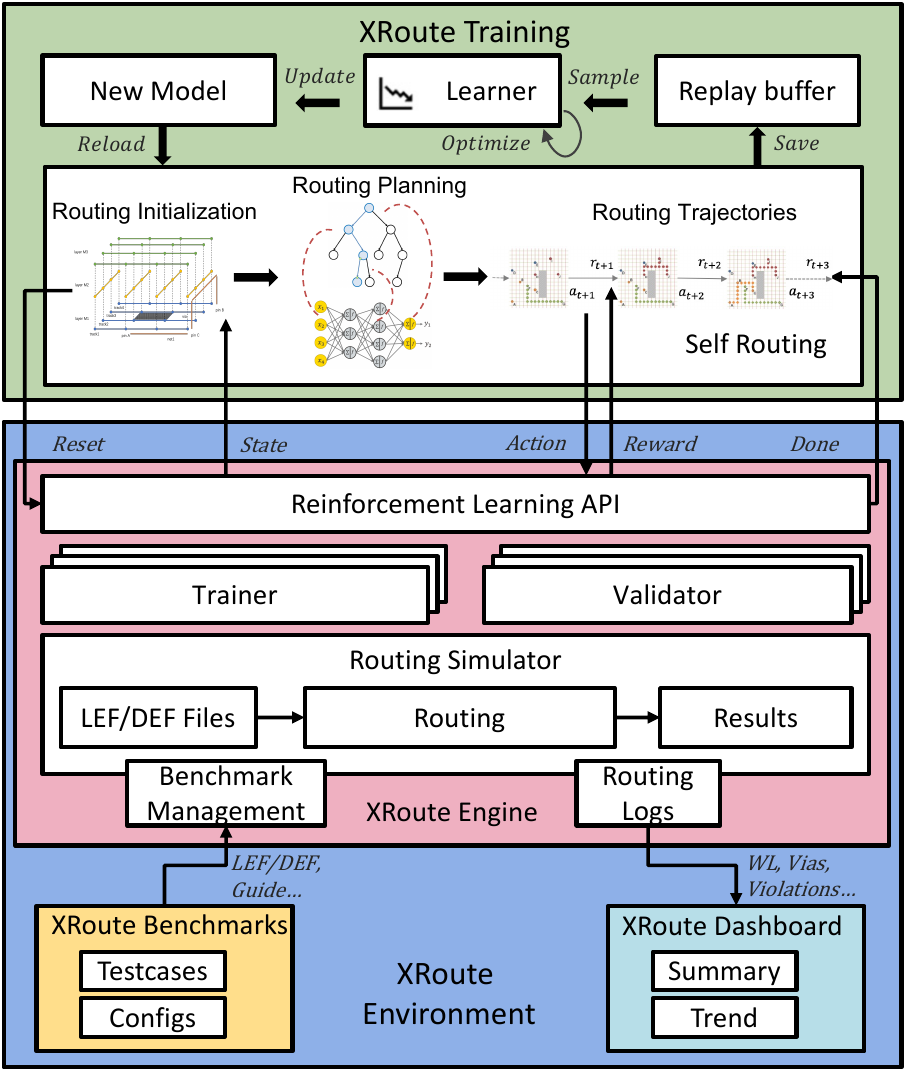}
    \caption{{\ours} architecture overview.}
    \label{fig:xroute_architecture}
\end{wrapfigure}

The overview of the {\ours} architecture is showed in Figure \ref{fig:xroute_architecture}. It is divided into two modules: {\ours} Environment and {\ours} Training. The {\ours} Environment is based on the {\ours} Engine, an advanced routing simulator built around a heavily customized version of the publicly available TritonRoute \citep{DBLP:journals/tcad/KahngWX21} \citep{Kahng2022TritonRoute}. The engine simulates a complete routing procedure from reading in the LEF, DEF and global routing guide files of a chip design, routing data preparation, region partition and initialization, net routing, design rules checking, rip-up and reroute, to outputting the final detailed routing results. Based on the routing simulator, {\ours} Engine provides two modes with similar API for RL-based algorithms' development. Trainer mode is used for model training. It can generate full-chip test beds of varying region size and dynamically movable routing regions with various layouts and netlists based on {\ours} Benchmarks. Different regions can be seen as different routing problems for RL-based algorithms and the trainer can define the number of training interactions in each region. According to the researcher's requirements, trainer mode can also define specific static regions with different pin density and number of nets for model training. Validator mode is used for model validation. It goes through all benchmarks with the well trained model and evaluate the routing results on wirelength, vias count and design rule violations. 

Both trainer mode and validator mode employ similar reinforcement learning APIs compatible with the OpenAI Gym API \citep{brockman2016openai} to interact with the {\ours} Training module. The interaction supports distributed deployment and multi-instance experiments. In distributed deployment, a ZeroMQ broker is created in between {\ours} Environment and {\ours} Training module for data transmission and protobuf mechanism is used for serializing structured data. 
{\ours} Benchmarks is a versatile set of benchmark tasks of varying size and difficulties that can be used to compare different algorithms, including ISPD-2018 and ISPD-2019 detailed routing testcases. A chip can be partitioned into regions with varying sizes, or user-defined width. RL-based algorithms is then trained and tested on this regions and the well trained model is validated by various benchmarks from {\ours} Benchmarks.
{\ours} Dashboard is another key component of {\ours} Environment, which collect the routing logs and results and present those data in a dashboard. It provides summary tables and trend graphs to show the variation of wirelength, vias used, design rule violations and many other routing evaluation indicators.

In {\ours} Environment, We design two RL-based tasks in our first version to improve the quality and efficient of detailed routing. One is net ordering, the other is net routing. They are the most crucial steps of detailed routing. We can follow the same way to define more routing RL tasks.

The {\ours} Training contains four components: self routing, replay buffer, learner and new model. The workers in self routing component read in the state of routing instances generated by the Routing Simulator, and make use of RL-based planning methods \cite{DBLP:conf/aaai/JinMJZCY22} to produce a routing action. The self routing worker pass the routing action to the Routing Simulator and get back the new state and routing cost, which is used to calculate the reward for the action. This procedure runs recurrently until the end of routing. Then a routing trajectory is generated and save in the replay buffer for later training in learner. Finally, new model generated by continuous training and optimizing are used for next self routing. The well trained model will be validated and compared with {\ours} benchmarks.

\subsection{{\ours} Engine}

In this section, two RL-based tasks of net ordering and net routing available in the first version of {\ours} Engine are presented. The detailed usage of these two tasks and the corresponding API and parameters are described in Appendix \ref{usage_of_xroute_environment}.

\subsubsection{Task 1: net ordering}


\paragraph{State and observations}
We define as state the complete set of data that is returned by the environment after actions are performed. On the other hand, we define as observation or representation any transformation of the state that is provided as input to the routing algorithms. At every step in the routing process, a RL agent receives an observation of the state of the routing region, which is viewed as a 3D grid graph. The grid graph is an essential part of detailed routing because the path search algorithm works directly on the grid graph, and various costs and properties are associated with the grid vertices and edges in the grid graph. Specifically, we follow \citep{DBLP:journals/tcad/KahngWX21} to partition the design into non-overlapping GCell-aligned regions, each of which has $C \times C$ GCells in one layer, where $C$ is a hyperparameter (e.g., $C = 3$, which means there are 9 GCells in each layer of the region). 
For each region, A nonregular-spaced 3D grid graph supporting irregular tracks and off-track routing is built. The terminal state is reached when all nets have been routed in the region.

While routing, we organize the environment state data as \emph{Dimension}, \emph{Nodes} and \emph{Nets} for each net.

\begin{itemize}
    \item \emph{Dimension}. The dimension of each region is represented by ($D_x$, $D_y$, $D_z$), which respectively indicate the number of lines (on-track line, off-track line, boundary line) in the vertical and horizontal directions of all routing layers, as well as the number of routing layers.
    \item \emph{Nodes}. The nodes are formed by all orthogonal lines, thus, there are $D_x \times D_y \times D_z$ nodes in the region. Each node is composed of 7 attributes: 1) maze index, the relative coordinates that all axes start from zero; 2) point, the absolute position of the node in the region; 3) type, being one of four values: \emph{NOTEXIST} if the intersection point is not exist in that layer, \emph{BLOCKAGE} for obstacle points, \emph{NORMAL} for normal line intersection points, and \emph{ACCESS} for access point \citep{DBLP:conf/dac/KahngWX20} of a certain net; 4) usage, a Boolean to indicate whether the node is used; 5) net, represents the net number to which the access point belongs; 6) pin, represents the pin number of the access point in the corresponding net; 7) cost, the edge cost in six direction (\emph{UP}, \emph{DOWN}, \emph{NORTH}, \emph{SOUTH}, \emph{EAST}, \emph{WEST}).
    \item \emph{Nets}. The set of the net numbers to be routed is designated as \emph{Nets}.
\end{itemize}

We propose a representation for the state in our baseline algorithms. We simply divide the points contained in the environmental state data into pin points, obstacle points, and empty points. When the environment is initialized, the pin point is the above-mentioned node of type=ACCESS. Obstacle points correspond to nodes of type=BLOCKAGE, and others are empty points. 
We set the access points and path points of prior routed nets into obstacle points before the next interaction. We construct the representation of the state from the perspectives of obstacles and nets in the 3D grid graph with dimension ($D_{x}$, $D_{y}$, $D_{z}$). For obstacles, we use a simple 0-1 feature. For nets, in addition to the basic 0-1 feature, we add six 0-1 features of each point to describe whether neighboring points belong to the same pin of the net. 
Our baseline model was trained with above representation, but researchers can easily do more feature mining and define their own representations based on the environment state. Here we provide some further references:

\begin{itemize}
    \item \emph{Scale feature}. The size of each net.
    \item \emph{Cross-layer feature}. The distribution of pins and access points of a net in the multi-layer routing environment.
    \item \emph{Boundary feature}. More resource conflicts and design rule violations in the boundary of the routing environment.
    \item \emph{Spatial feature}. The overlaps area of one net with other nets.
\end{itemize}

\ignore{<Add the state representation in DQN and PPO for net ordering here. For example, the size of the each net, the number of pins, the intersection of nets...>}

\paragraph{Action}
The action is represented by the number of a net chosen for subsequent routing, and the action space consists of all unrouted nets. At each step, {\ours} Engine will route the net that is selected by RL agent with a pre-defined path search routing algorithm.


\paragraph{Reward}
A well-performing routing solution is expected to exhibit reduced numbers of specified metrics such as wirelength, vias used, design rule violations. We define the reward function \ref{eq:net_ordering_reward_function} as below:
\begin{equation} \label{eq:net_ordering_reward_function}
    R(s, a, s') = cost(s) - cost(s')
\end{equation}
\begin{equation} \label{eq:net_ordering_cost_function}
    cost(s) = 0.5 \times wirelength + 4 \times via(s) + 500 \times drv(s)
\end{equation}

{\ours} Engine returns the total number of wirelength, vias used and design rules violations in state $s'$ after performing a action $a$ in state $s$. The cost function can be defined as a weighted sum of the increment of these metrics \citep{Liu2019ISPD}. Function \ref{eq:net_ordering_cost_function} is the cost function used in our baseline algorithms. We apply lower penalties to the agent for the action that results in smaller increments, ensuring that the agent learns a better net ordering strategy. It also allows researchers to add custom reward functions using wrappers which can be used to investigate reward shaping approaches.

\subsubsection{Task 2: net routing}


\paragraph{State and observations}
The state of routing environment in net routing include data of \emph{Dimension}, \emph{Nodes}, \emph{Nets} in net ordering task, with additional data as below:

\begin{itemize}
    \item \emph{Net}. The current routing net is represented by the serial number of the net.
    \item \emph{Head}. The current head of the routed path of the routing net.
\end{itemize}

The state representation can be defined as a 12-dimensional vector. The first three elements are the $(x, y, z)$ coordinates of the current agent position in the environment. The fourth through the sixth elements encode the distance in the $x$, $y$, and $z$ directions from the current agent location to the target pin location. The remaining six dimensions encode the usage or cost information of all the edges the agent is able to cross. This encoding scheme can be seen as a mix of the current state, the navigation, and the local capacity information. If the agent takes an illegal action or a connection to the final pin of the net from current state $s_t$, then the next state $s_{t+1}$ is the terminal state. Researchers can easily define their own representations based on the environment state by creating wrappers similarly.


\paragraph{Action}
The actions are represented with the coordinate of the start point $(x, y, z)$ defaulted to be the current head location, an integer $d$ from 0 to 5 corresponding to the direction of move from the start point, and an integer $s$ corresponding to the number to step to move, with a default value as 1.


\paragraph{Reward}
The reward is defined as a function of the selected action $a$, the current state $s$ and the next state $s'$, as shown in Function \ref{eq:net_routing_reward_function}.

\begin{equation} \label{eq:net_routing_reward_function}
    R(s,a,s') = 
    \begin{cases}
        HPWL(net) & \text{if all pins are connected}\\
        cost(s)-cost(s') & \text{otherwise}\\
    \end{cases}
\end{equation}

$HPWL(net)$ is the half-perimeter wirelength, the half-perimeter of the bounding box for all pins of the current routing $net$. This design encourages the agent to learn a path as short as possible since any unfruitful action will cause a decrement in the cumulative reward. Additionally, we limit the minimum total reward the agent can get when routing each net to be greater than a minimum threshold $T_{min}=-HPWL(net)*pin\_count(net)$. This scheme is a useful indicator to distinguish if the overall routing problem was successfully solved, or no feasible solution was found.

\subsection{{\ours} Benchmarks and Dashboard}


The {\ours} Engine is an end-to-end and flexible RL environment for detailed routing with many features that lets researchers try a broad range of new ideas. To facilitate fair comparisons of different algorithms and approaches in this environment, we also provide a set of pre-defined benchmark tasks that we call {\ours} Benchmarks. The routing results are shown on {\ours} Dashboard.

\paragraph{Benchmarks and metrics}

{\ours} Benchmarks are a versatile set of benchmark tasks of varying size and difficulties that can be used to compare different algorithms, based on ISPD-2018 \citep{Mantik2018ISPD} and ISPD-2019 \citep{Liu2019ISPD} detailed routing testcases in our first version. These two benchmark suites have totally 20 testcases in 65nm, 45nm, and 32nm technology nodes. The size is up to 899404 standard cells and 895253 nets. The ISPD-2019 benchmark suite includes more realistic routing rules compared to the ISPD-2018 benchmark suite, which makes the testcases more challenging and closer to industrial routing problems. Experiments are performed on these benchmark tasks with self-defined region size. When training, the environment can be dynamically moved to next region after pre-defined iterations.

\begin{figure}[!ht]
    \centering
    \includegraphics[width=1\textwidth]{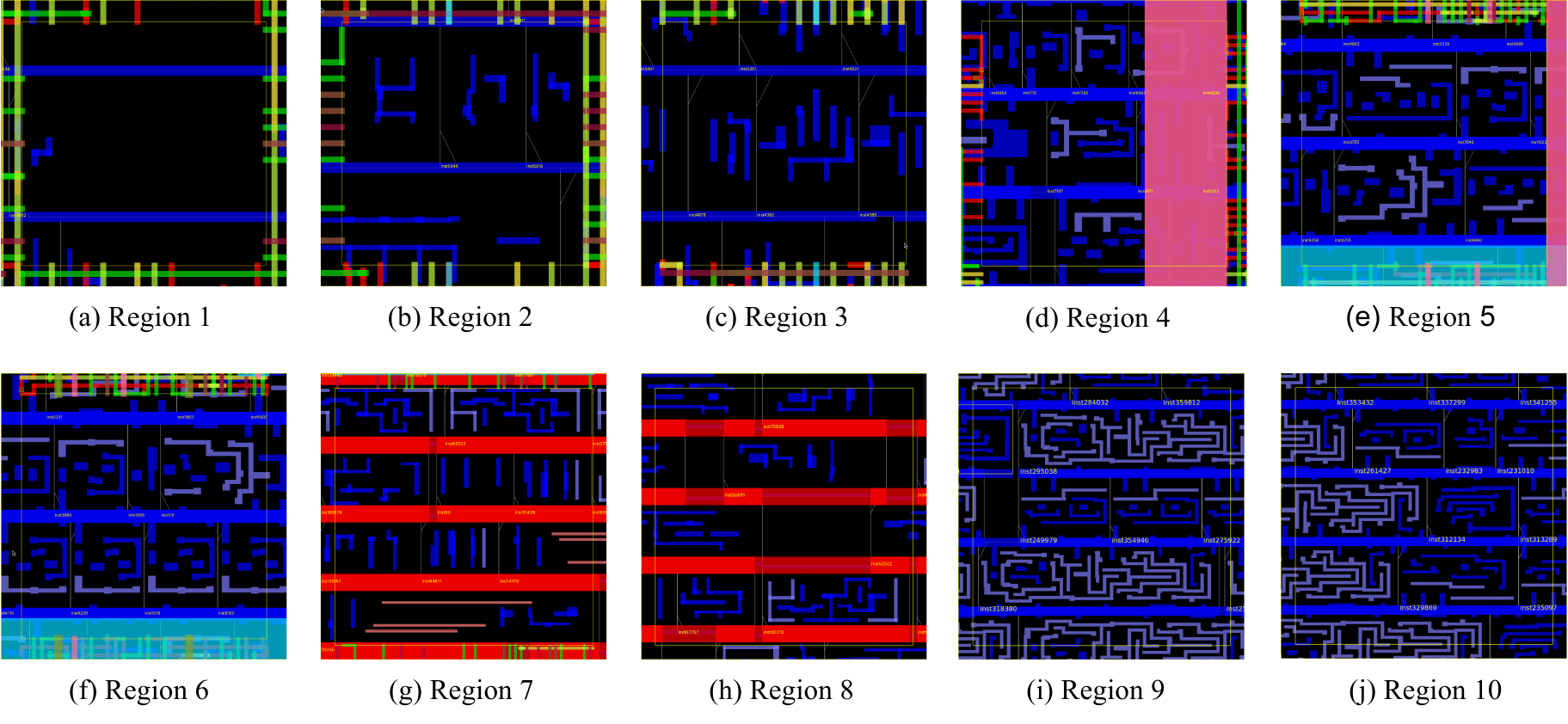}
    \caption{Detailed layout of pre-defined static regions.}
    \label{fig:static_regions_layout_detail}
\end{figure}

Training agents on dynamic regions in benchmark suite can be challenging. To allow researchers to quickly iterate on new research ideas, we provide several static regions with various size, pin density and number of nets for routing learning and testing. Learning converges faster in these regions. They can be considered as “unit tests” for RL-based algorithms where one can obtain reasonable results within minutes or hours instead of days or even weeks. Table \ref{tab:static_region_characteristics} shows the characteristics of the pre-defined static regions. We employ a method that involves dividing the total number of nodes in each region by the number of networks to be routed. This yields a value that describes the sparsity of each region, which serves as one of several metrics used to measure the difficulty of routing. Figure \ref{fig:static_regions_layout_detail} illustrates the layout details of the corresponding region through OpenROAD's debugging capabilities \citep{DBLP:conf/dac/AjayiCFHHKKLMNP19}. We also provide APIs to define new static region according to the requirements of researchers (See Appendix \ref{usage_of_xroute_environment}
for more details).

\begin{table}[!ht]
    \caption{Static region benchmarks characteristics}
    \label{tab:static_region_characteristics}
    \centering
    \scriptsize
    \begin{tabular}{@{}ccccccc@{}}
    \toprule
        Benchmark & From & \tabincell{c}{Size\\(\# GCells)} & \# Nets & \# Pins & Sparsity & \tabincell{c}{Position\\(Lower-left to upper-right)}\\
    \midrule
        Region1 & ISPD-2018 test1 & 1x1 & 36 & 30 & 1628.40 & (199500, 245100), (205200, 250800)\\
        Region2 & ISPD-2018 test1 & 1x1 & 49 & 51 & 1104.48 & (142500, 233700), (148200, 239400)\\
        Region3 & ISPD-2018 test1 & 1x1 & 49 & 66 & 394.88& (148200, 210900), (153900, 216600)\\
        Region4 & ISPD-2019 test3 & 2x2 & 58 & 75 & 3663.60 & (138000, 132000), (144000, 138000)\\
        Region5 & ISPD-2019 test3 & 2x2 & 67 & 101 & 2484.00 & (96000, 186000), (102000, 192000)\\
        Region6 & ISPD-2019 test3 & 2x2 & 73 & 122 & 1994.33 & (108000, 186000), (114000, 192000)\\
        Region7 & ISPD-2018 test5 & 3x3 & 63 & 124 & 1245.56 & (1566000, 639000), (1575000, 648000)\\
        Region8 & ISPD-2018 test5 & 3x3 & 55 & 127 & 846.82 & (1332000, 1224000), (1341000, 1233000)\\
        Region9 & ISPD-2019 test7 & 3x3 & 133 & 285 & 765.74 & (612000, 1314000), (621000, 1323000)\\
        Region10 & ISPD-2019 test7 & 3x3 & 137 & 281 & 651.61 & (648000, 1386000), (657000, 1395000)\\
    \bottomrule
    \end{tabular}
\end{table}

We evaluate two widely used RL algorithms on part of the {\ours} Benchmarks, providing a set of reference results for future comparison. We follow the ISPD-2019 detailed routing contest \citep{Liu2019ISPD} to evaluate the performance of different routers by total wirelength, vias used, design rule violations, and runtime.



\paragraph{Dashboard}

\begin{figure*}[!ht]
    \centering
    \subfigure[Summary table of routing results]{
        \begin{minipage}[t]{0.95\textwidth}
        \centering
	    {\includegraphics[width=1\textwidth]{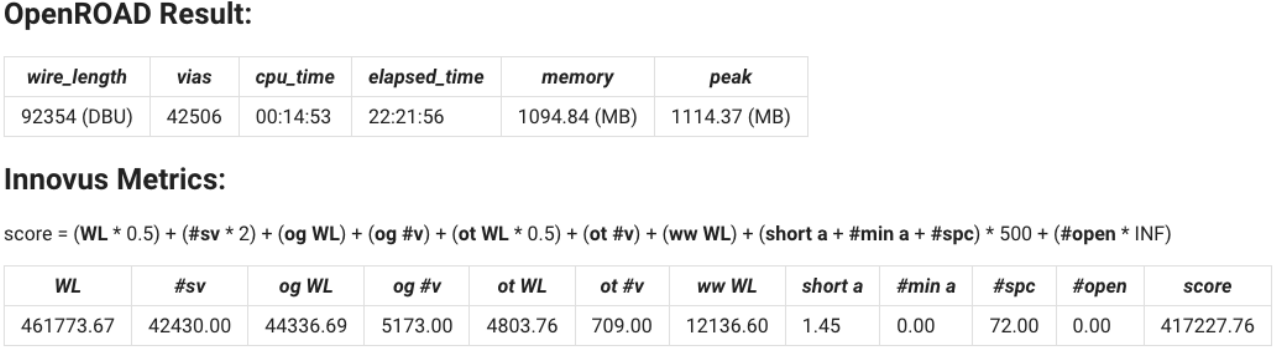}} 
	\end{minipage}
	\label{fig:dashborad_summary_table}
    }
    \subfigure[Trend graph of routing metrics]{
        \begin{minipage}[t]{0.95\textwidth}
        \centering
   		\includegraphics[width=1\textwidth]{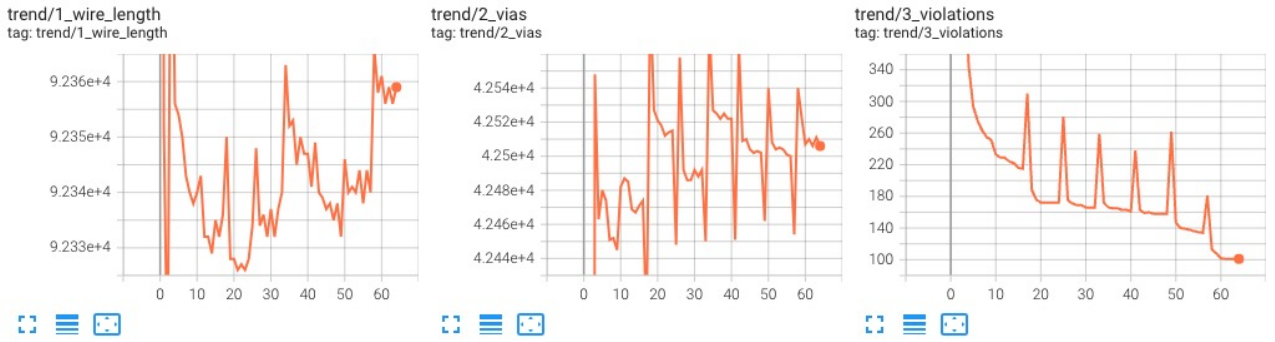}
        \end{minipage}
	\label{fig:dashboard_trend_graph}
    }
    \caption{A summary table and a trend graph in {\ours} Dashboard}
    \label{fig:xroute_dashboard}
\end{figure*}

{\ours} Dashboard is designed to visually present the detailed routing results such as wirelength, vias used, design rule violations and many other routing evaluation indicators in a dashboard during model validation. It collect the routing logs and results and present those data with summary tables and trend graphs as Figure \ref{fig:xroute_dashboard} shows. We also integrate the results from a commercial routing tools.

\section{Experiment and results}

\subsection{Environment setup and baseline}

We separate the {\ours} Environment and {\ours} Training (or RL-based agent) into different processes in a Linux server. Researchers can distribute these processes into separate servers. For net ordering task, A*-search algorithm is used for routing a net in Routing Simulator after the net is recommended by the RL agent. We partition the chip into regions with different sizes such as $1\times1$, $2\times2$, or $3\times3$ GCells. A 3D grid graph of the region is created based on the previous routed environment before routing each net in the region. We then route the nets on this grid during training and evaluation. 
During training, the hyperparameters (See Appendix \ref{usage_of_xroute_environment}
for more details) of net ordering task is defined and the network is trained by processing the same region several times. Then the training dynamically switch to a next region with different layouts and netlists in the same testcase. We use the final network checkpoint generated by the training process for evaluation.  We evaluate the agent on part of {\ours} Benchmarks. The baselines for comparison include TritonRoute, DQN \citep{mnih_human-level_2015} and PPO \citep{schulman2017proximal}. TritonRoute is the known-best academic routers for detailed routing, while DQN and PPO are widely used RL algorithms.

We conduct all experiments on a Linux server with a 40-core Intel Xeon CPU E5-2650 at 2.3 GHz and 240 GB shared memory, equipped with 2 NVIDIA GeForce GTX Titan XP GPU with totally 24 GB video memory.

\subsection{Agent architecture}

We modify the state transformation of PPO and DQN to address the challenge of inconsistent dimensions of the 3D grid state. We transform the original 3D grid state and each action to a fixed vector with $s$ dimensions (we use 64 in our experiments) by 3D convolution. 
For DQN implementation, we set the $Q$ function as a 2-layer fully connected network, with 128 and 64 hidden units in each layer. Each layer is followed by an exponential linear unit function. The input size of the $Q$ function is $s*2$, and the output size is 1. For actor-critic style PPO, our actor function has the same parameter architecture as the above $Q$ function. While the critic function consists of 2 fully connected layers with 64 hidden units in each layer. Each layer is followed by a hyperbolic tangent function. The input size of the critic function is $s$, and the output size is 1.


Detailed architecture and hyperparameters of DQN and PPO is presented in Appendix \ref{agent_architecture_and_hyperparameters}.

\subsection{Experimental results}

\begin{figure}[!ht]
    \centering
    \includegraphics[width=1\textwidth]{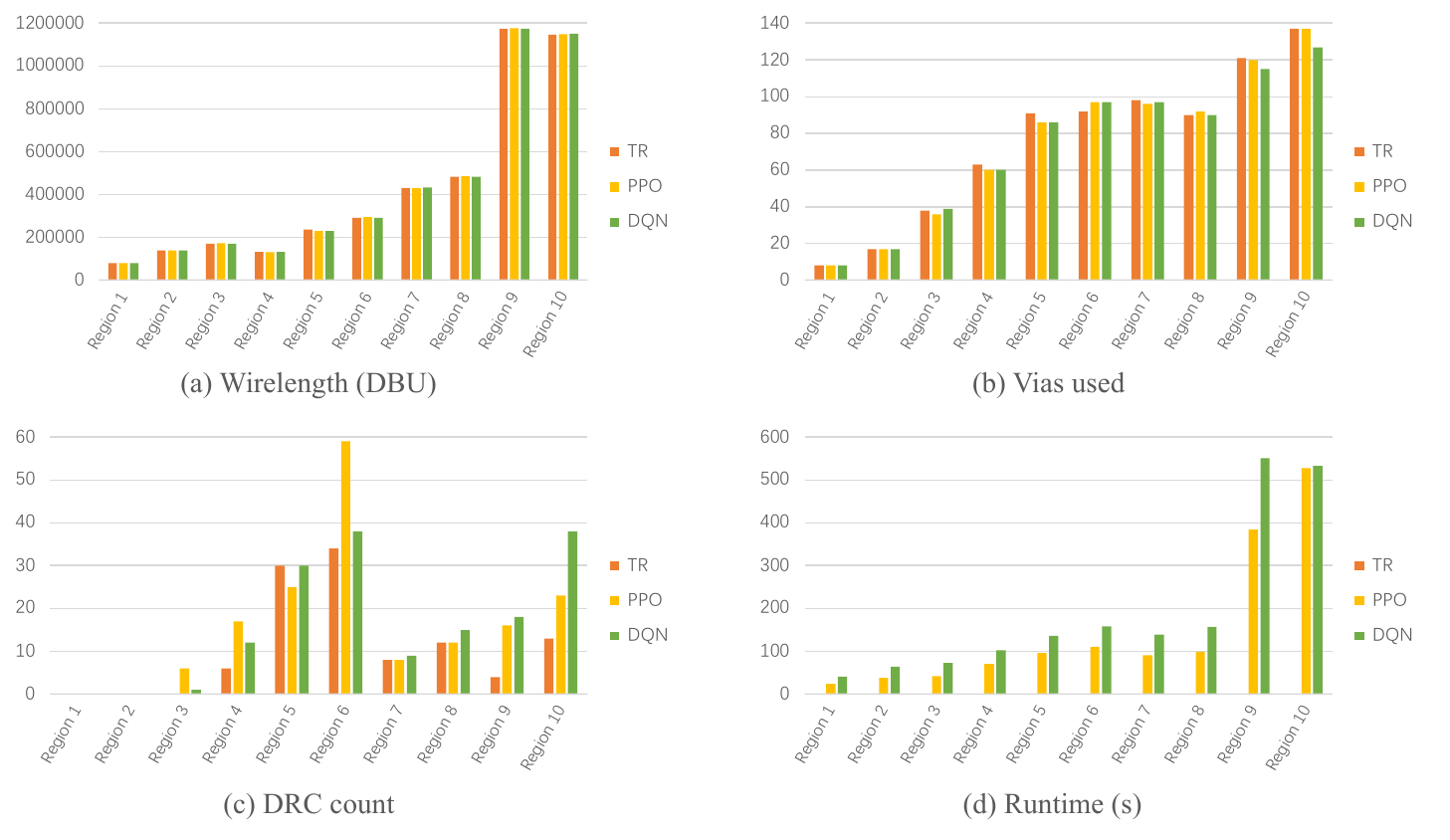}
    \caption{Comparison of metrics for ten static regions between TritonRoute (TR), PPO and DQN}
    \label{fig:static_regions_comparison}
\end{figure}

We trained the PPO and DQN model on every $1\times1$ GCell region of test1 in ISPD-2018 benchmark suite. There are totally 4556 different $1\times1$ GCell regions in the testcase and both of the self-routing process of PPO and DQN traversed all regions. After 4600 training episodes, we use the trained model to route ISPD-2018 test1 with region size of $1\times1$. We compare the performance of TritonRoute (with queue-based ripup and reroute in each region), PPO, and DQN on wirelength, vias used, design rule violations in Table \ref{tab:xroute_result_comparison}. We find that the well trained PPO has less design rule violations than TritonRoute after 40 routing iterations. We also use the trained model of PPO and DQN to route the 10 pre-defined static regions and the results are shown in Figure \ref{fig:static_regions_comparison}. It shows that TritonRoute (with queue-based ripup and reroute in static regions) performs best in most of the static regions. However, the well trained PPO in Region5 and Region7 and DQN in Region5 perform better for almost all metrics except the runtime. It means RL-based algorithms may be used in detailed routing for large scale chips if we have proper training environments. More experimental results are shown in Appendix \ref{additional_experimental_results}.


\begin{table*}[!ht]
    \caption{Comparison of metrics for ISPD-2018 test1 with region size of $1\times1$ GCell between TritonRoute (TR), PPO and DQN}
    \centering
    \scriptsize
    \medskip
    \begin{tabular}{@{}cccccccccccccccc@{}}
    \toprule
        \multirow{2}{*}[-0.7ex]{Iteration} & \multicolumn{3}{c}{Wirelength $(DBU)$} && \multicolumn{3}{c}{Vias used} && \multicolumn{3}{c}{DRV count} && \multicolumn{3}{c}{Runtime $(s)$}\\
    \cmidrule{2-4} \cmidrule{6-8} \cmidrule{10-12} \cmidrule{14-16}
        & TR & PPO & DQN && TR & PPO & DQN && TR & PPO & DQN && TR & PPO & DQN\\
    \midrule
        0th & 92689 & 92820 & 92842 && 41254 & 40372 & 40365 && 3162 & 5027 & 5057 && 28 & 1539 & 12806\\
        10th & 92378 & 92351 & 92356 && 42175 & 42691 & 42433 && 206 & 240 & 261 && 7 & 321 & 2029\\
        20th & 92344 & 92372 & 92357 && 42253 & 42792 & 42517 && 169 & 182 & 196 && 5 & 259 & 1707\\
        30th & 92347 & 92379 & 92369 && 42223 & 42801 & 42497 && 158 & 163 & 163 && 5 & 220 & 1478\\
        40th & 92346 & 92381 & 92370 && 42215 & 42792 & 42506 && 157 & 156 & 157 && 5 & 186 & 1330\\
        50th & 92349 & 92388 & 92375 && 42209 & 42829 & 42504 && 150 & 143 & 145 && 16 & 175 & 1258\\
        60th & 92353 & 92387 & 92378 && 42205 & 42800 & 42475 && 125 & 122 & 126 && 26 & 132 & 1026\\
        final & 92353 & 92387 & 92378 && 42205 & 42800 & 42475 && 125 & 121 & 126 && 26 & 125 & 993\\
    \bottomrule
    \end{tabular}
    \label{tab:xroute_result_comparison}
\end{table*}

\section{Conclusion and future work}

In this paper, we presented the {\ours} Environment, a novel open-source reinforcement learning environment for VLSI detailed routing. It is challenging, easy to use, customize and add additional scenarios, and it has specific functionality geared towards research in reinforcement learning. We provided the {\ours} Engine, a highly optimized C++ routing simulator; the {\ours} Benchmarks, a set of reference tasks to compare different reinforcement learning algorithms; the {\ours} Dashboard, a dashboard to present the detailed routing results such as wirelength, vias used and design rule violations during model validation. We expect that these components will be useful for investigating current scientific challenges in VLSI detailed routing. We present two tasks for learning to improve the quality and efficient of detailed routing. One is net ordering, the other is net routing. We can follow the same way to define new more routing reinforcement learning tasks. 

In the future, it would be interesting to investigate: 1) adding more RL tasks such as pin access analysis, design rules violation prediction and fixing, net ordering and net routing for global routing; 2) building more planning-based learning approaches \cite{DBLP:conf/aaai/ShenZXZP20,DBLP:journals/ai/ZhuoK17,DBLP:journals/ai/Zhuo014,DBLP:journals/ai/ZhuoM014,DBLP:journals/ai/ZhuoYHL10} to further help routing with novel planning techniques \cite{DBLP:journals/ai/JinZXWK22}; 3) adding more benchmarks from industrial chips and RL baseline algorithms for performance comparison; 4) accelerating the training and validating process in our routing environment.

\ignore{
\section*{Acknowledgement}
This research was funded by the National Natural Science Foundation of China (Grant No. 62076263), Guangdong Natural Science Funds for Distinguished Young Scholar (Grant No. 2017A030306028), Guangdong special branch plans young talent with scientific and technological innovation (Grant No. 2017TQ04X866), Pearl River Science and Technology New Star of Guangzhou and Guangdong Province Key Laboratory of Big Data Analysis and Processing.
}

\newpage
\bibliographystyle{plainnat}
\bibliography{references}

\newpage
\appendixautorefname
\appendix
\begin{wrapfigure}{r}{0.5\textwidth}
    \centering
    \includegraphics[width=0.48\textwidth]{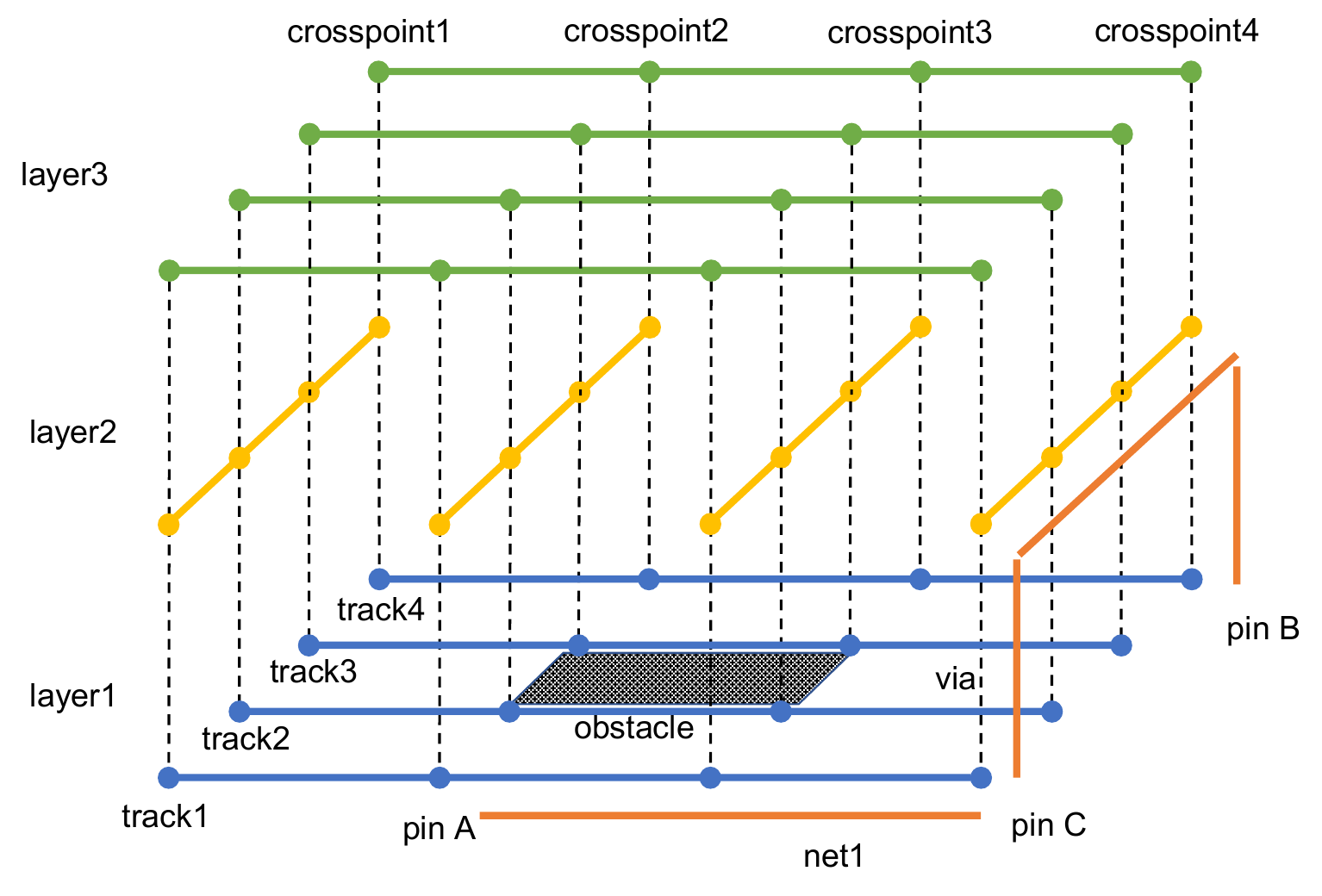}
    \caption{3D grid graph for detailed routing.}
    \label{fig:3d_grid_graph_for_routing}
\end{wrapfigure}

\section{Background of detailed routing}
\label{background_of_detailed_routing}

\subsection{Preliminaries}


\paragraph{Routing Space} VLSI routing is usually carried out after all circuit components 
are placed on the chip. Detailed routing ensures correct connection among those components and optimizes the connection paths to satisfy time budget, design rules, 
and many other physical, electrical, and manufacturing constraints. Routing is on a stack of metal layers, each of which has a preferred direction for routing, either horizontal or vertical. The preferred directions of adjacent layers are orthogonal to each other in order to minimize signal crosstalk. A wire segment routes along the preferred direction on the regularly spaced tracks, which are pre-defined according to the minimum width and spacing constraint of wire. Wires on adjacent metal layers can be electrically connected by vias in the crosspoint of tracks. 
On each track, there are a series of crosspoints viewed as vertices. The vertices and connections among them on all metal layers compose a 3D grid graph for detailed routing, as shown in Figure \ref{fig:3d_grid_graph_for_routing}. A vertex is uniquely defined by a 3D index $<l, t, c>$, which is a tuple of layer index, track index, and crosspoint index along the track. Adjacent vertices are connected by on-track wire segment edges on the same layer, or cross-layer via edges. Over the chip, there are some routing obstacles that vias and wire segments should avoid to prevent short and spacing violations. In Figure \ref{fig:3d_grid_graph_for_routing}, $net_1$ has three pins $(A, B, C)$ to be connected, and the orange paths are one of the connection schemes. Note that there can be as many as millions of nets to be routed in real chips.

\begin{figure*}[!ht]
    \centering
    \subfigure[Parallel run length spacing]{
        \begin{minipage}[t]{0.22\textwidth}
			\includegraphics[width=0.9\textwidth]{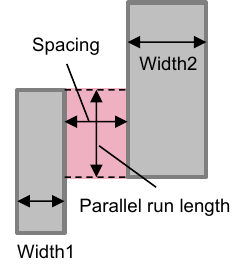} 
		\end{minipage}
		\label{fig:constraint_prl}
	}
    	\subfigure[End of line spacing]{
    		\begin{minipage}[t]{0.22\textwidth}
   		 	\includegraphics[width=0.9\textwidth]{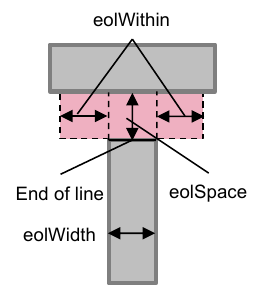}
    		\end{minipage}
		\label{fig:constraint_eol}
    	}
	\subfigure[Adjacent cut spacing]{
		\begin{minipage}[t]{0.22\textwidth}
			\includegraphics[width=0.9\textwidth]{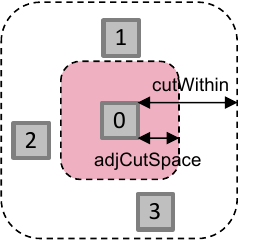} 
		\end{minipage}
		\label{fig:constraint_cut}
	}
    	\subfigure[Corner-to-corner spacing]{
    		\begin{minipage}[t]{0.22\textwidth}
		 	\includegraphics[width=0.9\textwidth]{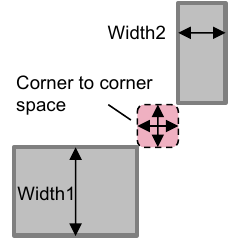}
    		\end{minipage}
		\label{fig:constraint_ctc}
    }
    \caption{Representative design rule constraints.}
    \label{fig:design_rule_constraints}
\end{figure*}

\paragraph{Constraints} The representative connectivity constraints and routing rules \citep{Liu2019ISPD} required to be satisfied by detailed routing are as shown below:

\begin{itemize}
    \item Open constraint: All pins of each net need to be fully connected. If any pin in a net is disconnected, the net will be considered as an open net.
    \item Short constraint: A via metal or wire metal of one net cannot overlap with via metal, wire metal, obstacle, or pin of another net, or the intersection part are the short area.
    \item Spacing rules: The spacing rules specifies the required spacing between two objects as shown in Figrue \ref{fig:design_rule_constraints}. Parallel run length (PRL) spacing in Figure \ref{fig:constraint_prl} defines the spacing requirement for two metal objects with PRL (i.e., the projection length between them). End of line (EOL) indicates that an edge that is shorter than eolWidth. EOL edge requires spacing greater than or equal to eolSpace beyond the EOL anywhere within eolWithin distance, as Figrue \ref{fig:constraint_eol} shows. Adjacent cut spacing in Figure \ref{fig:constraint_cut} specifies the minimum spacing allowed between via cuts that are less than cutWithin distance on the same net or different nets. Corner-to-corner (CtC) spacing specifies the required spacing between a convex corner and any edges. It is triggered when the parallel run length between two objects is less than or equal to 0, as shown in Figure \ref{fig:constraint_ctc}.
    \item Minimum area: All polygons on the layer must have an area that is greater than or equal to minimum area.
\end{itemize}

There are several preference metrics generally used to evaluate a detailed routing solution for better timing, routability and manufacturability, such as routing guide honoring, wrong-way routing, off-track routing et al.

\subsection{Routing problem}

The detailed routing problem can be formally defined as follows. Given a placed netlist, routing guides, routing grid graph, and design rules, detailed routing aims at successfully routing all nets by optimizing the weighted sum of total wire length, total vias used, nonpreferred usage (including out-of-guide, off-track wires/vias, and wrong-way wires), and design rule violations (including short, spacing, and minimum area violations). Note that design rule violations are highly discouraged and suffer much more significant penalty than others.

In detailed routing, sequential routing is widely adopted due to its scalability and flexibility. However, sequential routing is sensitive to the order of nets to be routed. Net ordering and net routing are two crucial parts in detailed routing. In the literature, the net orders are mostly determined by simple heuristic rules tuned for specific benchmarks, while A$^*$ search algorithm is used in most of routers for routing nets. We decouple those two key steps from TritonRoute \citep{Kahng2022TritonRoute} and create a RL-based environment to support utilizing reinforcement learning approaches to select the most appropriate net under a specific circumstance and route the nets efficiently and effectively, and then follow the convention of sequential routing for other nets. Essentially, nets are routed one after another, viewing previously routed nets as blockages. After all nets are routed in one round with possible violations, several rounds of rip-up and reroute iteration help to clean them up.

\begin{figure}[!ht]
    \centering
    \includegraphics[width=1\textwidth]{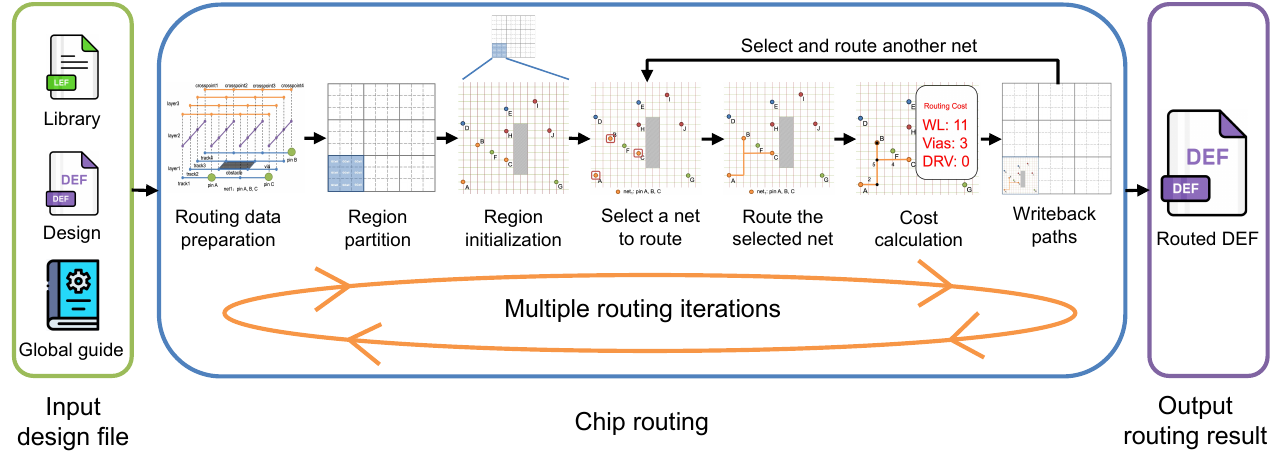}
    \caption{Detailed routing procedure.}
    \label{fig:detailed_routing_procedure}
\end{figure}

The detailed routing procedure is showed in Figure \ref{fig:detailed_routing_procedure}. After receiving the placement result, technical definition, design rules and other routing requirements, the router first analyzes the input data and initializes the parameters and chip environment before routing. Then multiple routing iterations are executed. In each iteration, the router adopts divide-and-conquer strategy to divide the entire chip into regions (each region may cover multiple GCells), finds the connection points between regions, and summarizes the final routing result after routing nets in each region separately. During region routing, the routing environment is initialized and the unrouted nets are selected and routed one by one according to the strategy of the routing algorithm. Multiple unrouted nets can be selected for parallel routing. If the routing space is projected to two dimensions, the layer assignment should also be considered to obtain the final three-dimensional routing solution. Detailed routing should satisfy the design rules and various constraints, as well as reducing the wirelength and the vias used. Net ordering and net routing are core steps in routing algorithms. After routing nets, the cost weighted by wirelength, vias and design rule violations is calculated. Violation areas and the affected nets are recorded and the violations are tried to be removed by rip-up and reroute. Finally, the resulting paths are written back to the chip routing space. Violations at region conjunctions can be handled by adding region offset in next iterations. After several rounds of iterations, the final routing results are presented and used for future design steps.

\section{Usage of {\ours} Environment}
\label{usage_of_xroute_environment}

In this section, we provide a brief introduction on utilizing the {\ours} Environment. For more detailed instructions, please refer to our GitHub repository \footnote{https://github.com/xplanlab/xroute\_env}.

\subsection{Parameters}

\subsubsection{Net ordering task}

{\ours} Environment provides a series of parameters that can be used to customize the behavior of the Routing Simulator. The following parameters are available:

\begin{itemize}
    \item \verb|testcase_name|: The parameter specifies the name of the testcase to execute routing. {\ours} Environment supports a total of 20 testcases from ISPD-2018 and ISPD-2019 in our first version.
    \item \verb|testcase_loop|: The number of times to run a testcase.
    \item \verb|region_size|: The size of the routing region that each router worker is responsible for. For instance, if the value is set to be 3, the design is then divided into a number of regions, each of which has $3 \times 3$ GCells. The router workers perform routing on each region separately.
    \item \verb|region_location|: The lower-left coordinate of a routing region within a design. If the parameter is not set, the router will perform routing on all regions sequentially.
    \item \verb|region_loop|: The number of times to run the router on a routing region.
    \item \verb|iteration_count|: The number of iterations for routing. It is common to perform multiple iterations during a routing of a chip design, where each iteration generates violations that will be resolved in the next iteration. This value ranges from 0 to 64.
    \item \verb|region_name|: The parameter specifies the static region to execute routing. 
    \item \verb|thread_count|: The number of workers running in parallel to expedite the routing process.
\end{itemize}

\subsubsection{Net routing task}

In addition to the parameters mentioned in the net ordering task, net routing has the following parameters:

\begin{itemize}
    \item \verb|net_loop|: The parameter represents the number of times to route each net in a region. The final result of each net is preserved, and the following routing is affected by  previous routed nets.
\end{itemize}

\subsection{API and sample usage}

To provide a better out-of-the-box experience, {\ours} Environment is designed to be compatible with the OpenAI Gym API \citep{brockman2016openai}. The code snippets shown below run random agents on our environment in training mode or evaluation mode for a static region or a complete testcase.

\subsubsection{Training mode}

The following code snippet initiates a training environment for ISPD-2018 test1 with region size of $1\times1$ GCell. Normally, the training will exit after completing routing without violations or after 65 iterations. To allow the agent to learn sufficiently, we set the \verb|testcase_loop| parameter to be 50, which means the agent will perform 50 routings on the chip design.

\lstset{
  framesep=10pt,
  xleftmargin=10pt,
  xrightmargin=10pt,
}
\begin{python}
from xroute.envs import ordering_training_env

env = ordering_training_env(testcase_name="ispd18_test1",
                            testcase_loop=50, region_size=1)
observation = env.reset()
done = False
while not done:
    action = env.action_space.sample()
    observation, reward, done = env.step(action)
\end{python}

\subsubsection{Evaluation mode}

The code snippet shown below launches an evaluation environment for ISPD-2019 test1 with region size of $2\times2$ GCells, and will exit after the 6th iteration is completed.

\lstset{
  framesep=10pt,
  xleftmargin=10pt,
  xrightmargin=10pt,
}
\begin{python}
from xroute.envs import ordering_evaluation_env

env = ordering_evaluation_env(testcase_name="ispd19_test1",
                              region_size=2, iteration_count=6)
observation = env.reset()
done = False
while not done:
    action = env.action_space.sample()
    observation, reward, done = env.step(action)
\end{python}

\subsubsection{Static region mode}

Users can utilize the provided codes below to customize their own static regions, within which routing can be performed.

\lstset{
  framesep=10pt,
  xleftmargin=10pt,
  xrightmargin=10pt,
}
\begin{python}
from xroute.utils import generate_static_region
from xroute.envs import static_region_env

generate_static_region(region_location=(36800, 228000),
                       save_dir="custom/Region1")

env = static_region_env(region_dir="custom/Region1")
observation = env.reset()
done = False
while not done:
    action = env.action_space.sample()
    observation, reward, done = env.step(action)
\end{python}

\section{Agent architecture and hyperparameters}
\label{agent_architecture_and_hyperparameters}

We describe the details of agent architecture and hyperparameters of PPO and DQN for net ordering task below.

\subsection{Model input}


In each reinforcement learning interaction, our {\ours} Environment provides the agent with environment state in the form of a list of nodes, which form a 3D grid of the routing region. The dimensions of those 3D grids are not fixed because of irregular tracks and off-track routing. Based on the state information returned by the environment, we construct a 3D obstacle grid representing environmental obstacle information and multiple 3D net grids representing net features as the input of the model.

Obstacle grid describes the feature of obstacle points. If a point is an obstacle point, the value is 1, otherwise it is 0. Empty points or pin points occupied in previous routing interactions are processed as obstacle points in new interactions with the environment.

In our net ordering task, all nets that need to be routed constitute the action space for reinforcement learning. Obviously, the size of this action space is also not fixed. Assume that there are $N$ initial nets as optional actions, there will only be $N-1$ optional actions after a net is routed in the next interaction. In addition, the number of initial optional actions for each net ordering task is different. Net grid describes the feature of distributed pins of a net. In addition to the feature of whether the grid point is a pin point, net grid also presents whether the adjacent points in the six directions belong to the same pin. For example, if the current grid point is a pin point, and its east adjacent point is also a pin point belonging to the same pin, then the current grid point is assigned a value of 1, otherwise it is 0. The feature extraction of adjacent points in other directions can be followed in the same way. Therefore, the net grid contains 7 feature channels.

\subsection{Model outputs}
Assume that the routing region has $N$ unrouted nets, the current action space contains $N$ actions, numbered as $0, 1, ... , N-1$ respectively. The model scores each action, then converts them into a probability distribution of nets, and return the action number with the highest probability as the output.

\subsection{Baseline model structure}
We use PPO \citep{schulman2017proximal} and DQN \citep{mnih_human-level_2015} as the baseline algorithms to evaluate the routing performance and results.

\subsubsection{PPO model structure}

\begin{figure}[!ht]
    \centering
    \includegraphics[width=0.9\textwidth]{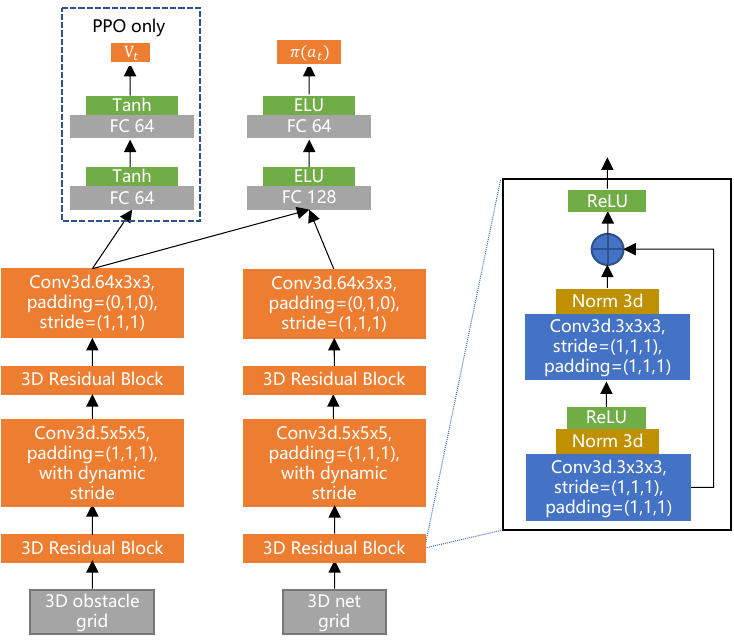}
    \caption{Model structure of PPO and DQN.}
    \label{fig:baseline_model_structure}
\end{figure}

The model structure of PPO is shown in Figure \ref{fig:baseline_model_structure}. Due to the challenge of dimensional inconsistency of the 3D grid state, we make appropriate modifications to the state transitions of PPO. We define two consecutive 3D convolutional layers and 3D batchnorm layers as a 3D residual block. The kernel of the 3D convolution is $3\times3\times3$, the stride is 1, and the padding is 1. Overall, we achieve the conversion from the original obstacle grid and net grid to 64-dimensional fixed vectors in three steps.

At first, we use 3D residual blocks to extract higher feature matrix without changing the size. At the second step, we align the input 3D grid to the standard size of $64\times64\times3$. We convolve the matrix with a 3D convolutional layer with kernel size 5, stride 1, and padding 1 to ensure that the output size is smaller than or equal to the standard size. During the convolution process here, we dynamically set the step size according to the following formula \ref{stride}. Then we pad the output matrix with zeros to make it in accord with the standard size. At the third step, we further use a 3D residual block to extract high level features. The size of the feature matrix remains $64\times64\times3$. We then use a 3D convolution with kernel size = (64,3,3), stride=(1,1,1), padding=(0,1,0) to convert the dimension of the feature matrix into (1,64,1), and further convert it into a 64-dimensional vector.

\begin{equation}
    stride = \lceil (input\_size-standard\_size)/standard\_size \rceil + 1\label{stride}
\end{equation}


After the obstacle grid is converted into a 64-dimensional vector, it is passed to a value network to obtain the state value. As shown in the dashed box in Figure \ref{fig:baseline_model_structure}, the value network is a 2-layer fully connected network with 64 hidden units in each layer, followed by Hyperbolic Tangent (Tanh) function.

Finally, we concatenate the two 64-dimensional vectors together, and pass the concatenated vectors to a policy network to calculate the score of each net. The policy network is a 2-layer fully connected network with 128 and 64 hidden units in each layer, followed by an exponential linear unit (ELU) function. Then we convert the scores of all nets into the probability distribution of each net, and return the action number with the highest probability as the output.

\subsubsection{DQN model structure}
 The model structure of DQN is the same as PPO in Figure \ref{fig:baseline_model_structure} except the network components in the dashed box are removed.

\subsection{Hyperparameters}
The detailed hyperparameters of the PPO and DQN model are shown in Table \ref{hyper:PPO} and Table \ref{hyper:DQN}. We give more descriptions for the following parameters.


\begin{itemize}
    \item \verb|Maximum Training Episodes|: We train the PPO and DQN model on every $1\times1$ GCell region of test1 in ISPD-2018 benchmark suite. Each GCell has different number of nets that need to be routed, and an episode is generated after routing a GCell. This hyperparameter indicates the number of GCells traversed by our agent during training.
    \item \verb|Model Update Period|: During training, this hyperparameter defines the updates frequency of the target network, with episode used as the unit of measurement.
\end{itemize}

\ignore{
DQN for task 2: The Q-network consists of three fully connected layers with 32, 64, and 32 hidden units in each layer. Each layer is followed by a ReLU (Rectified Linear Unit) activation layer. The input size is 12, which is the same as the number of elements in the designed state vector. The output size is 6, aligned with the number of possible next states.
}

\begin{table}[!ht]
    \centering
    \small
    \begin{minipage}[t]{0.4\textwidth}
        \caption{Hyperparameters of PPO}
        \centering
        \begin{tabular}{lr}
            \toprule
                \textbf{Parameter} & \textbf{Value}\\
            \midrule
                Maximum Training Episodes & 4600\\
                Model Update Period & 100\\
                Batch Size & 8\\                
                Optimizer & Adam\\
                Discount Factor & 0.99\\     
                Observation dimension & 64\\                         
                Number of Actors & 1\\
                Actor Learning Rate & 0.0003\\
                Critic Learning Rate & 0.001\\      
                Clipping $\epsilon$ & 0.2\\
                Training Epochs per Update & 10\\                
            \bottomrule
        \end{tabular}
        \label{hyper:PPO}        
    \end{minipage}
    \makebox[0.08\textwidth]{}
    \begin{minipage}[t]{0.4\textwidth}
        \caption{Hyperparameters of DQN}
        \centering
        \begin{tabular}{lr}
            \toprule
                \textbf{Parameter} & \textbf{Value}\\
            \midrule
                Maximum Training Episodes & 4600\\
                Model Update Period & 300\\                  
                Batch Size & 8\\   
                Optimizer & Adam\\     
                Discount Factor & 0.99\\      
                Observation dimension & 64\\   
                Number of Actors & 1\\                
                Learning Rate & 0.002\\                
                Capacity of Replay Buffer & 1000\\
                Greedy Factor& 0.9\\
            \bottomrule
        \end{tabular}
        \label{hyper:DQN}
    \end{minipage}
    
\end{table}

\section{Additional experimental results}
\label{additional_experimental_results}

In addition to the experimental results of ISPD-2018 test1 with region size of $1 \times 1$ GCell
, we have also conducted experiment with region size of $3 \times 3$ GCells on the same test case. The performances of different algorithms are shown in Table \ref{tab:xroute_result_comparison_18t1_3x3}.

\begin{table*}[!ht]
    \caption{Comparison of metrics for ISPD-2018 test1 with region size of $3\times3$ GCells between TritonRoute (TR), PPO and DQN}
    \centering
    \scriptsize
    \medskip
    \begin{tabular}{@{}cccccccccccccccc@{}}
    \toprule
        \multirow{2}{*}[-0.7ex]{Iteration} & \multicolumn{3}{c}{Wirelength $(DBU)$} && \multicolumn{3}{c}{Vias used} && \multicolumn{3}{c}{DRV count} && \multicolumn{3}{c}{Runtime $(s)$}\\
    \cmidrule{2-4} \cmidrule{6-8} \cmidrule{10-12} \cmidrule{14-16}
        & TR & PPO & DQN && TR & PPO & DQN && TR & PPO & DQN && TR & PPO & DQN\\
    \midrule
        0th & 89499 & 89607 & 89580 && 37600 & 37241 & 37275 && 4695 & 5616 & 5544 && 4 & 6387 & 4723\\
        10th & 89149 & 88724 & 88731 && 37753 & 37772 & 37638 && 21 & 29 & 36 && 4 & 1167 & 1487\\
        20th & 89143 & 88742 & 88746 && 37764 & 37790 & 37728 && 19 & 23 & 24 && 2 & 984 & 1205\\
        30th & 89140 & 88732 & 88742 && 37761 & 37794 & 37723 && 15 & 17 & 18 && 2 & 1620 & 1172\\
        40th & 89143 & 88738 & 88743 && 37765 & 37792 & 37719 && 14 & 18 & 18 && 2 & 1624 & 1125\\
        50th & 89142 & 88739 & 88745 && 37763 & 37776 & 37718 && 14 & 19 & 18 && 10 & 1622 & 1111\\
        60th & 89142 & 88736 & 88745 && 37759 & 37777 & 37714 && 11 & 17 & 18 && 14 & 1607 & 1055\\
        final & 89142 & 88736 & 88745 && 37759 & 37777 & 37713 && 11 & 17 & 18 && 14 & 1642 & 1129\\
    \bottomrule
    \end{tabular}
    \label{tab:xroute_result_comparison_18t1_3x3}
\end{table*}

The results show that the region size can also affect the performance of algorithms. Compared to the results obtained with the region size of $1 \times 1$ GCell, the wirelength and vias used decrease significantly with the region size of $3 \times 3$ GCells, and it also leads to a faster reduction in design rule violations. This is because a larger routing region size provides more exploration space for the router to optimize its routing strategy, thus, researchers should choose an appropriate size for their experiments.



\end{document}